\documentclass[fleqn,11pt]{wlscirep}
\usepackage[utf8]{inputenc}
\usepackage[T1]{fontenc}
\usepackage{lineno}

\usepackage{booktabs}
\usepackage{multirow}
\usepackage{longtable}
\usepackage{xr-hyper}
\usepackage[charter,cal]{mathdesign}
\usepackage{ulem}
\usepackage{graphicx}
\usepackage{caption}

\usepackage{xspace} 

\makeatletter
\newcommand*{\addFileDependency}[1]{
  \typeout{(#1)}
  \@addtofilelist{#1}
  \IfFileExists{#1}{}{\typeout{No file #1.}}
}
\makeatother

\newcommand{\name}{{CoCoGraph}\xspace}

\title{
A collaborative constrained graph diffusion model for the generation of realistic synthetic molecules
}

\author[a]{Manuel Ruiz-Botella}
\author[a,*]{Marta Sales-Pardo}
\author[a,b,*]{Roger Guimer\`a}

\affil[a]{Department of Chemical Engineering, Universitat Rovira i Virgili, 43007 Tarragona, Catalonia}
\affil[b]{ICREA, 08007 Barcelona, Catalonia}

\affil[*]{Corresponding authors: Marta Sales-Pardo (E-mail: marta.sales@urv.cat); Roger Guimer\`a (E-mail: roger.guimera@urv.cat)}

\begin{abstract}
Developing new molecular compounds is crucial to address pressing challenges, from health to environmental sustainability. However, exploring the molecular space to discover new molecules is difficult due to the vastness of the space. Here we introduce \name, a collaborative and constrained graph diffusion model capable of generating molecules that are guaranteed to be chemically valid. Thanks to the constraints built into the model and to the collaborative mechanism, \name outperforms state-of-the-art approaches on standard benchmarks while requiring up to an order of magnitude fewer parameters. Analysis of 36 chemical properties also demonstrates that \name generates molecules with distributions more closely matching real molecules than current models. Leveraging the model's efficiency, we created a database of 8.2M million synthetically generated molecules and conducted a Turing-like test with organic chemistry experts to further assess the plausibility of the generated molecules, and potential biases and limitations of \name.

\end{abstract}
\begin{document}

\flushbottom
\maketitle



\section*{Main}

Discovering and developing novel molecular compounds is key to address several pressing challenges. These include the extensive and costly process of developing new drugs\cite{alakhdar2024}, the creation of advanced materials\cite{Menon2022},  the design of more environmentally friendly refrigerants\cite{Alkhatib2022}, the identification of unknown metabolites\cite{Young2024}, and the discovery of molecules that bind to disease-associated target proteins\cite{Ackloo2022}, among others. However, the vast molecular space of chemistry, estimated to comprise around \(10^{60}\) different molecules\cite{Polishchuk2013}, renders the discovery of new compounds a high-dimensional problem that is exceedingly complex. Consequently, these areas stand to benefit greatly from artificial intelligence systems capable of generating novel molecules with desired properties or of reconstructing molecules based on available molecular information.

Traditionally, algorithms for molecule generation have relied on rule-based models and optimization\cite{Kutchukian2010, Butler2018, Blaschke2020}. However, these classical approaches are limited to modifying existing molecules rather than generating entirely new ones \cite{Bilodeau2022}. With the progression of deep learning, new generative models for molecules have been developed employing techniques such as variational autoencoders \cite{jin2019}, generative adversarial networks \cite{decao2022} and graph neural networks (GNN) \cite{Mercado2020} . Although these models were groundbreaking and improved performance considerably, they still face challenges related to scalability, computational efficiency, molecular validity, and adherence to chemical constraints\cite{dai2020, liao2020, you2018}. Moreover, even the most advanced models within these categories exhibit limited generalization capabilities, often struggling to generate molecules that deviate significantly from those seen during training\cite{lee2023}. 

Advances in probabilistic diffusion models\cite{sohldickstein2015,ho2020,song2021,yang23} have led to innovative generative algorithmic techniques that alleviate some of these shortcomings, originally in image generation \cite{ramesh2021,ramesh2022} and later in other areas.
%
%
In the context of molecule generation \cite{Austin2021, hoogeboom2022, jo2022}, diffusion models involve a process that progressively adds noise (atoms and/or bonds) to a molecular graph, followed by a denoising process that learns to reconstruct molecules by removing the noise. The denoising process is then used to
%
generate new molecules. 
%
%
To better handle molecular constraints and improve sampling efficiency, graph-based diffusion models\cite{haefeli2023,qin2024} such as  DiGress \cite{vignac2023} and CDGS \cite{huang2023} 
employ discrete noise processes. 
Theoretically, this structure facilitates the generation of valid molecules and has the ability to produce novel molecules not seen during training. Nevertheless, creating generative models that accurately reflect the original molecular distributions while also generalizing to new molecules remains a challenge\cite{you2018,Yang2024,du2022}.
Therefore, further improvements in discrete diffusion techniques are necessary to ensure efficient generation of chemically valid molecules and, at the same time, the comprehensive exploration of the chemical space.

\begin{figure}[htbp]
  \centering
  \includegraphics[height=0.85\textheight]{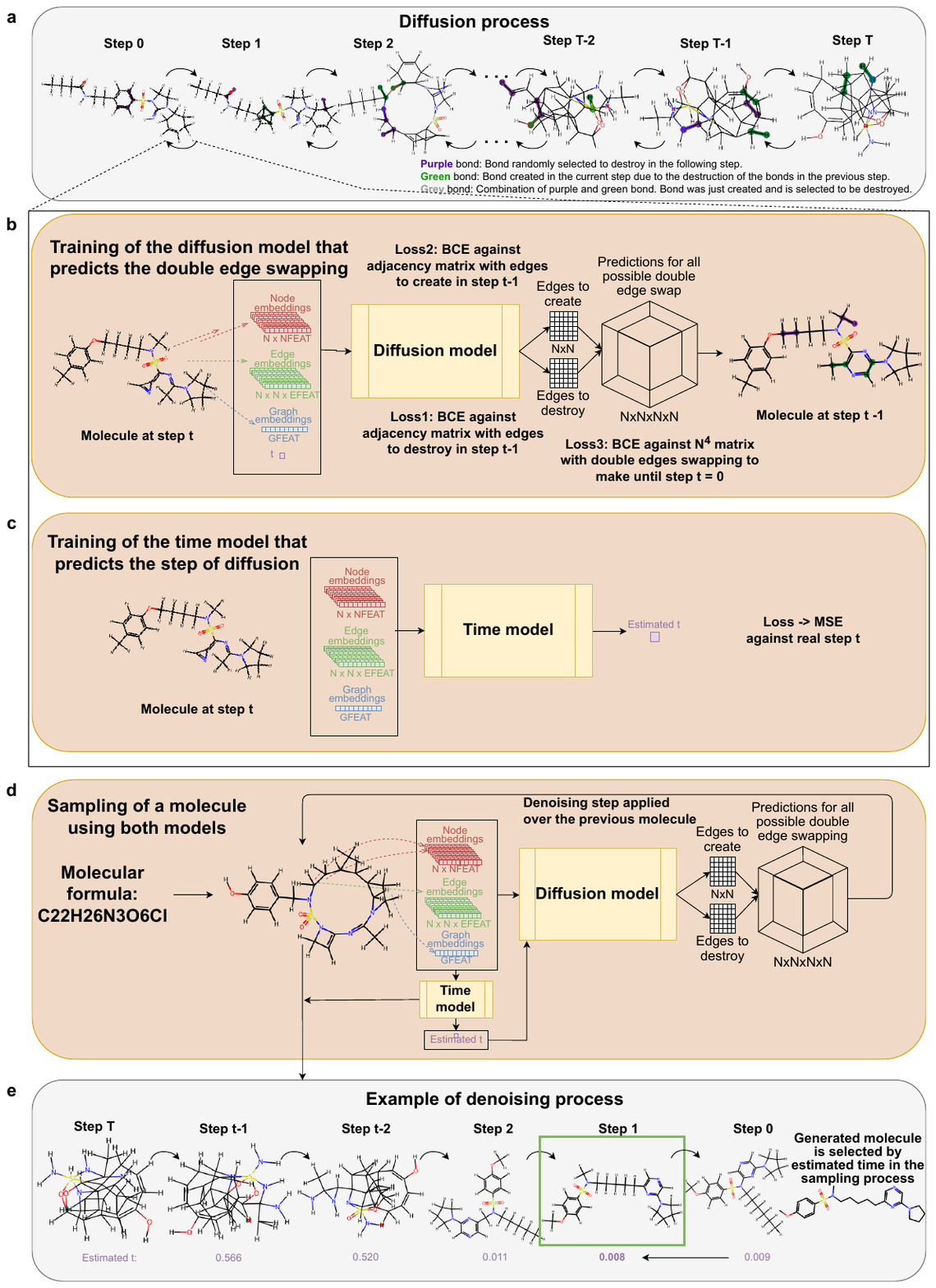} 
  \caption{{\bf Constrained collaborative graph diffusion model, \name.} {\bf a,} Constrained diffusion process. We introduce noise in the molecular graph by swapping two chemical bonds at each step.
  We then train diffusion and time models to revert this process. 
{\bf b,} Diffusion model. At each step, it receives molecular features and the timestep as an input and assigns a score to all possibilities of edge swaps.
{\bf c,} Time model. It receives molecular features and estimates the time step of the current molecular graph.
{\bf d,} Sampling. We use trained diffusion and time models in collaboration to generate a trajectory of denoising starting from a random molecular graph with a defined molecular formula. We then select the molecule with the smallest predicted time as the generated molecule.
  }
  \label{fig:model}
\end{figure}

With this goal in mind, here we introduce a collaborative constrained discrete diffusion model (\name) capable of generating molecules that are guaranteed to be chemically valid, that are very diverse, and whose chemical properties have distributions that closely resemble those of the known chemical universe. To achieve this, we introduce two key mechanisms into our diffusion model (Fig.~\ref{fig:model}). First, we use a discrete process that involves double edge swapping and constrains each atom to always have the correct valence~\cite{maslov02,milo04,kharel2022}, maintaining other chemical properties such as molecular weight, number of atoms, number of bonds, and molecular formula. Second, we introduce a collaborative mechanism in which two models are trained at each step of the denoising process. The first model ({\it diffusion model}) is trained to predict the double edge swapping operation to be reverted at each denoising step. This model takes as input molecular graph features and the diffusion time step, and outputs the probability for each possible double edge swap to be applied to the molecule. Additionally, we train a second model ({\it time model}), which learns to predict the time step of the diffusion process and {\it collaborates} with the diffusion model by informing it of how close the molecular graph is to the original molecule, so that the diffusion model can adapt its double edge swapping predictions to the actual (as opposed to expected) progress of the denoising process. 


Through this discrete, constrained, and collaborative approach, \name achieves 100\% chemical validity of novel generated molecules and beats the state-of-the-art on the most comprehensive existing benchmark. Additionally, an in-depth analysis of the distribution of tens of relevant molecular properties shows that our model generates molecules whose properties are statistically closer to real ones than those produced by current diffusion models.
%
%
Importantly, thanks to the constrains in the diffusion process and to the synergy between models, our approach achieves these results with an order of magnitude fewer parameters than existing models. The lightness of the model allows us to create a dataset with 8.2 millions of synthetically generated molecular structures, over which we conduct a Turing-like test to determine whether human experts can distinguish between generated molecules and original ones. We find that chemists with undergraduate or graduate knowledge of organic chemistry can only identify the real molecules with 62\% accuracy (59\% for those without masters' or PhD-level education), close to the 50\% random guess baseline. For specific groups of generated molecules, such as acyclic molecules or molecules with predominantly aliphatic bonds, expert performance is statistically compatible with 50\% accuracy, thus indicating the high quality of the generated molecules.
%
Our results underscore the effectiveness of our constrained collaborative model in navigating the vast chemical universe, %
and highlight the potential of our approach for real-world applications.

\section*{Results}

\subsection*{A collaborative constrained diffusion model for the generation of graphs with fixed degree sequence}

Although diffusion models for molecule generation initially used continuous noise distributions, discrete graph-based diffusion models such as DiGress \cite{vignac2023} and CDGS \cite{huang2023} have been shown to be superior. More recent models like Construct\cite{madeira2024g} have introduced constraint-aware diffusion processes specifically designed for graph generation. By using a noising diffusion process that is aware of some chemical constraints, and automatically satisfies them, these models are able to enforce specific chemical rules and properties during the generation process\cite{fishman2024, fishman2023}.

Here, we constrain even further the molecular graphs considered during a given diffusion process, so that each (noising and denoising) step preserves the nodes (atoms and, thus, molecular formula) and degree sequence (exact number of bonds per atom, that is, valence) \cite{maslov02,milo04,kharel2022}. To achieve this, at each noising diffusion step we swap two edges, so that two bonds $AB$ and $CD$ within the molecule are randomly selected and removed, and two new bonds $AC$ and $BD$ are formed 
(Fig.~\ref{fig:model}). By doing this, molecular graphs diffuse into a Molloy-Reed distribution~\cite{molloy95,maslov02,milo04}, which is the maximum-entropy distribution over the space of graphs with fixed degree sequence.
The satisfaction of chemical constraints by construction implies that: (i) invalid molecules not satisfying the constraints are never generated; (ii) the molecular structure search space is vastly reduced, because all structures not satisfying the constraints are automatically excluded; (iii) the chemical constraints do not need to be learned; and therefore (iv) models can be much smaller and focus on learning more subtle structural features of molecules.  

%
The denoising diffusion model learns to undo these edge swaps (Fig.~\ref{fig:model}; Methods). It takes as input the time step $t$ of the diffusion process and the molecular graph, which runs through three graph neural layers. The resulting node embeddings, together with edge features, molecular graph features and time, are used to estimate the plausibility of each edge swap by means of a feed forward network.

When using only the diffusion model, we observe that the progress of the diffusion process is not uniform across the training set---even after scaling the number of steps by molecular graph size (number of bonds), some molecules are quickly randomized whereas for others it takes much longer. Therefore, the time feature in the diffusion model turns out to be not very informative. Based on this observation, we introduce a time model (Fig.~\ref{fig:model}; Methods) that estimates how far the molecular graph is from the original molecule. This model takes as input the molecular graph and returns a normalized time, which is fed during sampling into the diffusion model instead of the actual time step, thus collaborating with the diffusion model by providing more relevant information about the actual position within the diffusion trajectory. Furthermore, at the conclusion of the sampling process, the model chooses the molecule with the smallest predicted time throughout the whole trajectory (that is, in principle, the closest to the original molecule), rather than the last generated molecule. The architecture of the time model is very similar to that of the diffusion model---the input graph goes through three graph neural layers, and the produced embeddings are used to predict time (Fig.~\ref{fig:model}).

A key consequence of our collaborative constrained diffusion approach, which we name \name, is a significantly more efficient model architecture that requires an order of magnitude fewer parameters than state-of-the-art approaches (Table~\ref{tab:model_comparison}). By incorporating chemical constraints directly into the diffusion process, our model inherently preserves chemical validity rather than needing to learn these rules. This reduction in model complexity translates into substantially lower computational requirements, making molecule generation more accessible. Importantly, our design based on constrains enables the model to allocate its learning capacity into capturing structural patterns of real molecules, resulting in better performance despite its smaller size, as demonstrated in the following sections.



\subsection*{\name outperforms existing generative models for molecules on a standard benchmark}

To comprehensively evaluate the performance of \name, we compare it to state-of-the-art molecular generative models using the GuacaMol \cite{Brown_2019} benchmark suite. This evaluation framework provides standardized metrics to assess the quality and diversity of generated molecules. We compared CoCoGraph to the juntion tree variational autoencoder (JTVAE) \cite{jin2019} and DiGress \cite{vignac2023}, with each model evaluated on their respective training datasets to ensure fair comparison.
We consider two different \name models (Table~\ref{tab:model_comparison}; Methods), both of which are smaller in number of parameters than the comparators. The BASE version of our model requires only 534K parameters in total (471K for the diffusion model and 63K for the time model), an order of magnitude fewer than  DiGress (4.6M) and JTVAE (5.3M). The enhanced FPS \name model, which incorporates molecular fingerprints as additional inputs to improve edge swapping prediction in the diffusion model and time prediction in the time model, uses 4.4M parameters.
\begin{table}[t!]
    \centering
    \begin{tabular}{lccccc}
    \hline
    \textit{Model} & \# \textit{parameters} & \textit{Validity (\%)} & \textit{Uniqueness (\%)} & \textit{Novelty (\%)} & \textit{KL divergence (\%)} \\ \hline
    JTVAE & 5.3M & 100 & 99.9 & 99.3 & 47.3  \\ 
    DiGress & 4.6M & 85.2 & 100 & 99.9 & 92.6  \\ 
    CoCoGraph (BASE) & 0.471M+0.063M & 100 & 99.9 & 98.6 & 96.0  \\ 
    CoCoGraph (FPS) & 3.1M+1.3M & 100 & 99.9 & 98.5 & 96.7 \\ \hline
    \end{tabular}
   \caption{{\bf Model comparison on the GuacaMol benchmark.} We show the number of parameters, the percentage of valid molecules (validity), the percentage of uniquely generated molecules ( uniqueness), the percentage of molecules not in the known chemical universe (novelty), and the KL divergence score for the BASE and FPS versions of \name, compared to JTVAE and DiGress. }
    \label{tab:model_comparison}
\end{table}
Both \name models achieve 100\% chemical validity for generated molecules, which is a direct consequence of the constrained diffusion approach inherently obeying chemical rules throughout the diffusion process. Importantly, perfect validity is achieved without sacrificing uniqueness (99.9\%) or novelty (98.5\% and 98.6\%) of the generated molecules, demonstrating that the constraints do not overly restrict the exploration of the chemical space. Benchmark algorithms JTVAE and DiGress also generate molecules that are virtually guaranteed to be unique and novel, although for DiGress this applies only to the 85.2\% valid molecules generated.

Beyond the general requirements of validity, uniqueness and novelty, the distribution of physicochemical properties for generated molecules is the critical metric for evaluating generative models. Indeed, we aim to generate molecules that are novel but plausible, that is, that have physicochemical properties that are statistically similar to those of real molecules. The GuacaMol benchmark quantifies this by measuring the Kullback-Liebler (KL) divergence between the distributions of ten physicochemical properties of generated molecules and those of real molecules.
As demonstrated by the KL divergence scores (Table~\ref{tab:model_comparison}), \name generates molecules whose property distributions more closely match those of real molecules, achieving scores of 96.0\% and 96.7\% for the BASE and FPS versions respectively, compared to 92.6\% for DiGress and 47.3\% for JTVAE. These scores correspond to typical KL divergences of 0.033 and 0.041 for \name models, 0.077 for DiGress, and 0.749 for JTVAE, so, even though the score difference does not seem very large, \name reduces the KL divergence by, on average, approximately a factor of two with respect to the best performing benchmark model, DiGress.
This indicates that our model captures the underlying distribution of molecular properties more accurately, producing molecules that better reflect the characteristics of the known chemical universe.

\begin{figure}[t]
  \centering
  \includegraphics[width=1\textwidth]{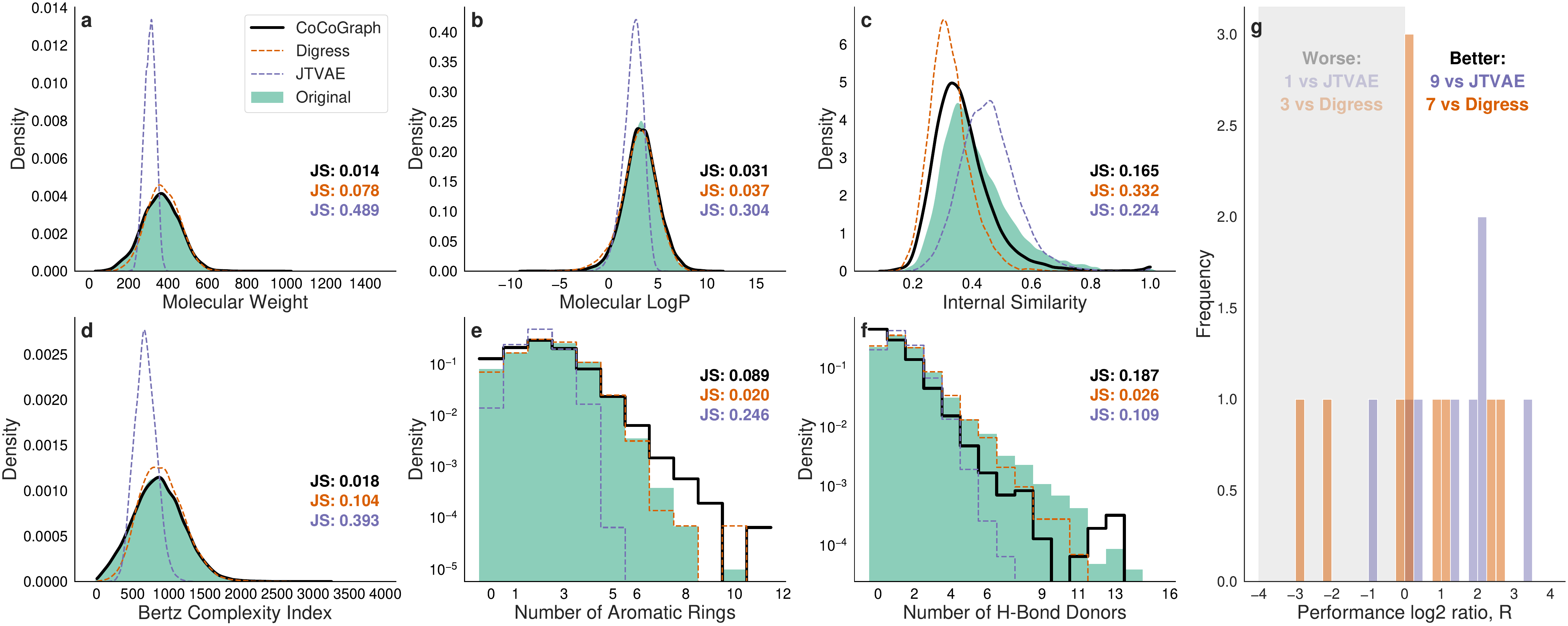} 
  \caption{{\bf Performance comparison on GuacaMol benchmark properties.} {\bf a-f}, Distributions of six molecular properties: {\bf a}, molecular weight; {\bf b}, molecular logP; {\bf c}, internal similarity; {\bf d}, Bertz complexity index; {\bf e}, number of aromatic rings; and {\bf f}, number of H-bond donors. For each property, the distribution of values calculated for molecules generated by \name  (black line) is compared to that of the original molecules (green distribution), and to those of molecules generated by JTVAE (purple dashed line) and DiGress (orange dashed line). Jensen-Shannon (JS) distance values between each model and the original distribution are shown. {\bf g}, Summary comparison based on the log2 ratio of JS distances between \name and comparator models for the properties in ({\bf a-f}). Positive values indicate \name outperforms the comparator model and vice versa.}
  \label{fig:guacamol}
\end{figure}

To provide more nuance into the KL divergence scores, we analyze in detail each of the ten specific molecular properties used in the GuacaMol benchmark. For this analysis, we focus on the FPS \name model (Fig.~\ref{fig:guacamol}; see Fig.~\ref{fig:combined_basemodel} for the BASE model). The distributions of properties like molecular weight, molecular logP, internal similarity, or Bertz complexity index are best approximated by \name (Fig.~\ref{fig:guacamol}A-D), as measured by the Jensen-Shannon (JS) distance. 
For a few other properties, such as the number of aromatic rings  (Fig.~\ref{fig:guacamol}E), \name outperforms JTVAE but shows lower performance than DiGress. Finally, for the number of H-bond donors both benchmark models achieve slightly better distribution matching than \name (Fig.~\ref{fig:guacamol}F).
All in all, \name FPS outperforms JTVAE on 9 out of 10 properties and DiGress on 7 out of 10 properties (Fig.~\ref{fig:guacamol}G). The BASE \name model performs slightly worse than the FPS model, but still better than both comparators (Fig.~\ref{fig:combined_basemodel}).

This improvement across models and multiple molecular characteristics underscores the effectiveness of our constrained collaborative approach in generating chemically valid and structurally novel molecules that are both realistic and diverse in terms of their physicochemical properties. The comparison to JTVAE is particularly illuminating in this regard. Indeed, despite the fact that JTVAE-generated molecules are always valid and plausible, the analysis of the distributions shows that their physicochemical properties are restricted to narrow ranges, which indicates that they are not as diverse as real molecules. The same, although to a much lesser extent, is generally true for molecules generated by DiGress, which is apparent from the fact that the mode of the distributions is typically higher for DiGress than for \name or real molecules.


\subsection*{Molecules generated by \name are plausible on a wide range of chemical properties}

While standard benchmarks like GuacaMol provide a useful starting point for evaluating molecular generative models, they only assess performance on a limited set of physicochemical properties. Additionally, as soon as a benchmark becomes standard, it starts being used during algorithm design and evaluation, potentially leading to overfitting of the corresponding properties. To provide a more comprehensive evaluation of our model's ability to generate chemically plausible molecules, we extended our analysis (after all models had been fully trained) to a diverse set of 36 chemical properties.
To avoid selection bias, we employed OpenAI's O1-mini model to identify a representative and diverse set of molecular descriptors that can be calculated with RDKit\cite{Greg2025}. This approach ensures that we consider properties deemed important by an external agent rather than selecting features where our model might excel. The properties span a wide range of molecular characteristics including size and composition metrics, topological features, electronic properties, and drug-likeness indicators (Table~\ref{tab:chemical_properties}).
\begin{table}[t!]
\centering
\begin{tabular}{lll}
\hline
\multicolumn{3}{c}{\textbf{Physicochemical properties}} \\
\hline
\\
\multicolumn{3}{l}{\textit{Basic Physicochemical Properties}} \\
\hline
Molecular Weight & Exact Molecular Weight & Heavy Atom Count \\
Number of Valence Electrons & N-H/OH Count & N-O Count \\
Fraction Csp3 & Quantitative Estimate of Drug-likeness & Balaban's J Index \\
\hline
\\
\multicolumn{3}{l}{\textit{Lipinski's Rule of Five Descriptors}} \\
\hline
Number of H-bond Donors & Number of H-bond Acceptors & Molecular LogP \\
Number of Rotatable Bonds & Topological Polar Surface Area & \\
\hline
\\
\multicolumn{3}{l}{\textit{Ring and Aromaticity Descriptors}} \\
\hline
Number of Aromatic Rings & Number of Aliphatic Rings & Ring Count \\
Number of Saturated Rings & Bertz Complexity & \\
\hline
\\
\multicolumn{3}{l}{\textit{Electronic Descriptors}} \\
\hline
Molar Refractivity & Maximum Partial Charge & Minimum Partial Charge \\
Maximum Absolute Partial Charge & Minimum Absolute Partial Charge & IPC \\
EState VSA Descriptor 1 & & \\
\hline
\\
\multicolumn{3}{l}{\textit{Topological Descriptors (Chi Descriptors)}} \\
\hline
Chi0 Index & Chi1 Index & Chi2n Index \\
Chi3n Index & Chi0n Index & \\
\hline
\\
\multicolumn{3}{l}{\textit{Van der Waals Surface Area (VSA) Descriptors}} \\
\hline
SlogP VSA Descriptor 1 & SlogP VSA Descriptor 2 & SlogP VSA Descriptor 3 \\
SlogP VSA Descriptor 4 & SlogP VSA Descriptor 5 & \\
\hline
\end{tabular}
\caption{{\bf The 36 chemical properties used to evaluate molecules generated by each model.} The properties were selected to identify diverse molecular properties subdivided in 5 groups that represent interesting molecular characteristics desired for molecule generation.}
\label{tab:chemical_properties}
\end{table}

\begin{figure}[t!]
  \centering
  \includegraphics[width=1\textwidth]{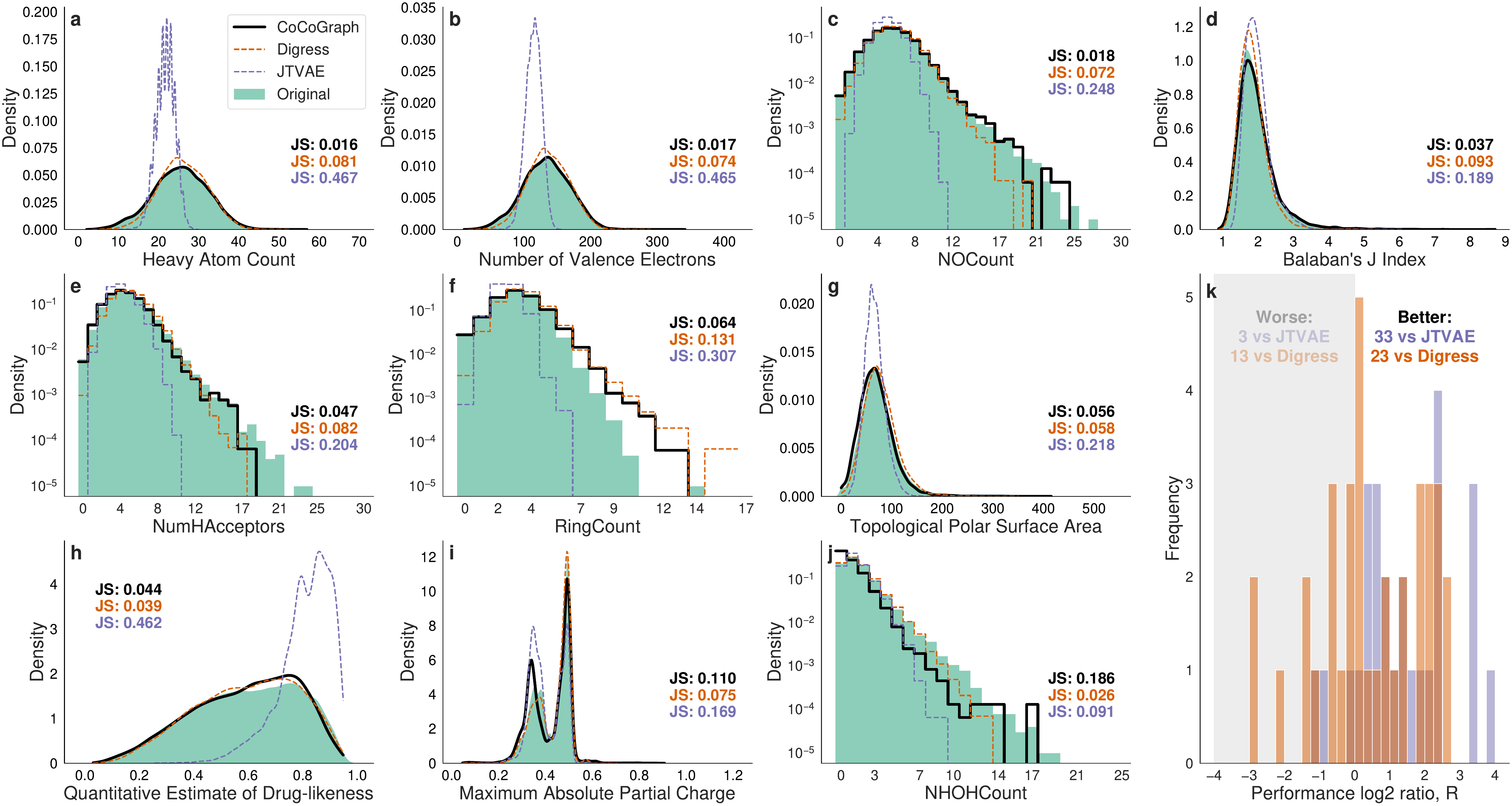} 
  \caption{{\bf Detailed performance comparison on a subset of 36 chemical properties.} {\bf a-j}, Distributions of ten molecular properties: {\bf a}, heavy atom count; {\bf b},  number of valence electrons; {\bf c},  NOCount; {\bf d},  Balaban’s J Index; {\bf e},  number of H acceptors; {\bf f},  ring count; {\bf g},  topological polar surface area (TPSA); {\bf h},  quantitative estimate of drug-likeness (QED); {\bf i},  maximum absolute partial charge; and {\bf j},  NHOHCount. For each property, the distributions for molecules generated by the CocoGraph FPS model (black line) is compared to that of the original molecules (green distribution) and to those of molecules generated by JTVAE (purple line) and DiGress (orange line). {\bf k},  log2 ratio of JS distances between CocoGraph FPS and the other models, where a positive value indicates that CocoGraph FPS outperforms the comparative model.}
  \label{fig:properties_distribution}
\end{figure}

Following the same methodology described in the previous section, we calculated JS distances between the distributions of each property for original molecules and molecules generated by different models (Table \ref{tab:js_supplementary}). In Fig.~\ref{fig:properties_distribution}, we show the distributions of ten properties: heavy atom count, number of valence electrons, NO count, Balaban's J index, number of H acceptors, ring count, topological polar surface area (TPSA), quantitative estimate of drug-likeness (QED), maximum absolute partial charge, and NHOH count.

The comprehensive comparison across all 36 properties is summarized in Fig.~\ref{fig:properties_distribution}k, which shows the $\log_2$ ratio of JS distances between models. \name outperforms DiGress on 23 out of 36 properties (64\%) and JTVAE on 33 out of 36 properties (92\%). The ten properties shown in Fig.~\ref{fig:properties_distribution}a-j were selected to reflect this performance distribution: the first seven properties (panels a-g) show \name outperforming both competing models; panels h and i show properties where CoCoGraph outperforms JTVAE but not DiGress; and panel j shows NHOH count, where both comparator models outperform \name. \name shows particular strength in topological features (Balaban's J index), electronic properties (number of valence electrons), and drug-likeness indicators (QED)---properties that are critical for applications in medicinal chemistry and drug discovery. 
In Fig.~\ref{fig:combined_basemodel_all}, we show that the BASE \name model also outperforms, overall, the benchmark algorithms, despite its much smaller number of parameters.
These results further validate the effectiveness of our constrained collaborative approach. 

\subsection*{A large database of realistic synthetic molecules}

The computational efficiency of \name, with its reduced parameter count, enables molecule generation at scale with modest computational resources. Our model produces thousands of chemically valid molecules per hour on a single mid-range GPU, allowing us to create a comprehensive database  containing 8.2 million molecules, with only 7.1\% redundancy. This high efficiency, combined with the 98.5\% novelty rate demonstrated in Table \ref{tab:model_comparison}, means our database contains approximately 7.6 million novel and unique, chemically valid molecules that were not present in the training data (see  Fig.~\ref{fig:molecule_grid} for a random sample of 50 generated molecules). 
Such a large-scale database of novel molecules may be a valuable resource for exploring new regions of chemical space. Given that the estimated molecular universe comprises approximately $10^{60}$ different molecules\cite{Polishchuk2013}, a systematic exploration of this extensive space remains a formidable challenge. Our synthetic molecule database offers a diverse collection of chemically valid structures that could accelerate discovery across multiple domains, including drug development, materials science, and catalysis research. 

To evaluate how realistic our synthetic molecules appear to domain experts, we developed a molecular Turing-like test. 
In this test, experts in organic chemistry, biochemistry, and related fields are presented with pairs of molecules sharing the same molecular formula---one real molecule from our original dataset and one generated by \name. Participants are asked to identify which molecule is original and which one is synthetic, and then repeat this exercise across 20 rounds (Fig.~\ref{fig:turing_test_web}). Participants also provide information about their level of expertise (university undergraduate or graduate training in organic chemistry).

\begin{figure}[t!]
    \centering
    \includegraphics[width=.9\textwidth]{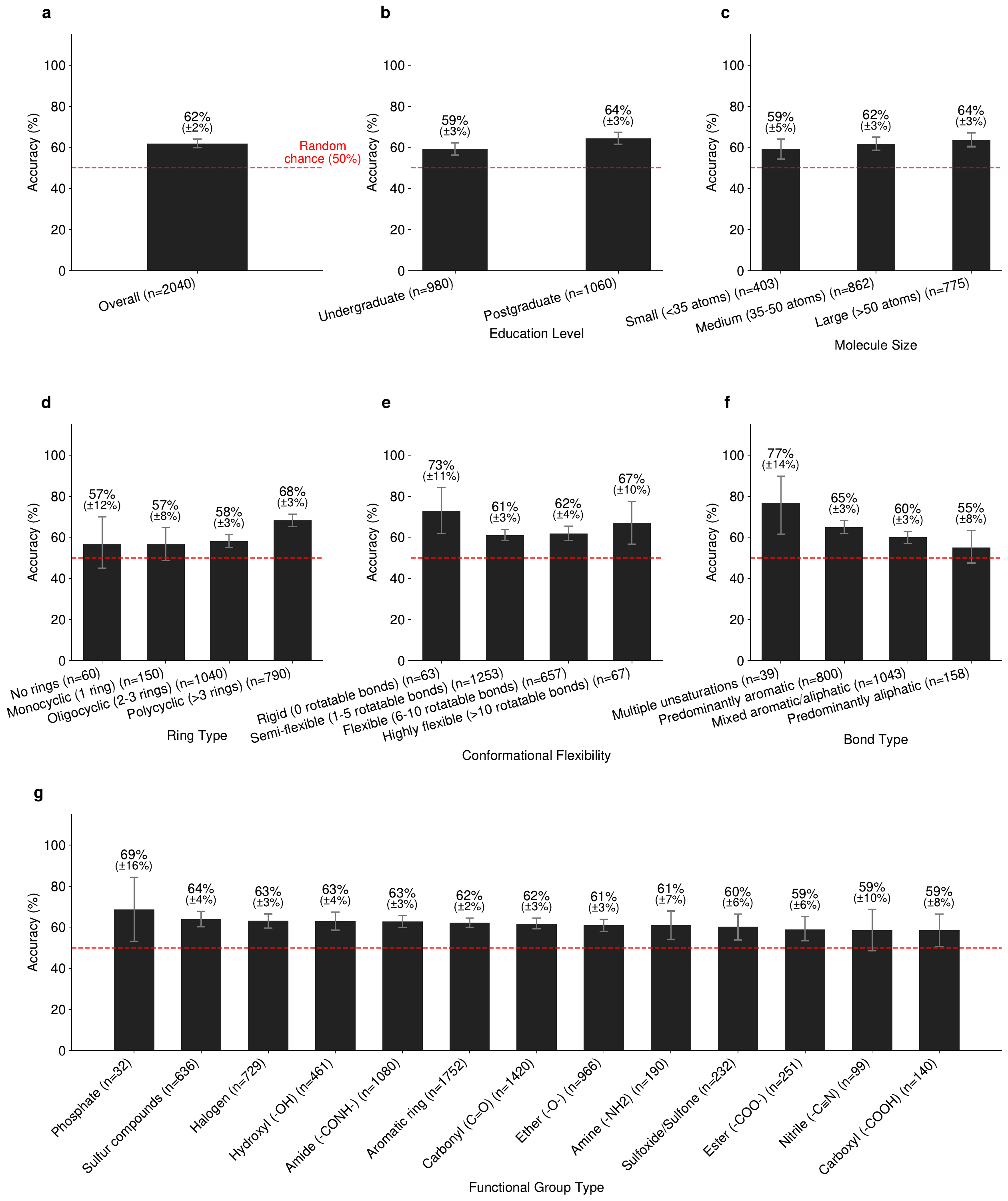} 
    \caption{{\bf Performance in the Turing-like test.} We assess the performance of participants in the Turing-like test by computing their accuracy at correctly identifying the original, non-generated molecule over all attempts. Error bars represent the standard error of the mean calculated via bootstrapping. {\bf a},  Overall accuracy of participants in the Turing-like test. {\bf b},  Accuracy by level of education in organic chemistry. {\bf c},  Accuracy by molecular size in terms of the number fo atoms. {\bf d},  Accuracy by ring type. {\bf e},  Accuracy by conformational flexibility of the molecules. {\bf f},  Accuracy by bond type. {\bf g},  Accuracy by functional group type. }
    \label{fig:turing_test}
  \end{figure}

We collected responses from 102 experts, totaling 2040 individual molecule pair assessments. The results reveal that experts achieved an overall AUC of 62\% at distinguishing real molecules from those generated by \name (Fig.~\ref{fig:turing_test}a). This accuracy is close to the 50\% baseline that would be expected from random guessing. Breaking down the results by level of expertise (Fig.~\ref{fig:turing_test}b) shows a slight improvement in discrimination ability with increased expertise---undergraduate participants achieved 59\% accuracy, while graduate participants achieved 64\% accuracy.

To further understand the strengths and potential biases of \name, we break down the analysis of the Turing-like test results along several dimensions (Fig.~\ref{fig:turing_test}c-g). In terms of the number of atoms, we observe that larger synthetic molecules are slightly easier for experts to identify (Fig.~\ref{fig:turing_test}c). This suggests that the difficulty of the generation task increases with the complexity of the molecule, although the difference is small. We also find that, in general, molecules with fewer rings are harder to classify for experts (that is, they are more reliably generated by \name), to the point that for acyclic molecules expert performance is compatible with random guessing (Fig.~\ref{fig:turing_test}d). Similarly, experts have more difficulty classifying molecules that are predominantly aliphatic than molecules that are predominantly aromatic or with multiple unsaturations (Fig.~\ref{fig:turing_test}f). In fact, for predominantly aliphatic molecules performance is again compatible with random guessing.
We find no clear tendency in the dependency on conformational flexibility (Fig.~\ref{fig:turing_test}e), or significant differences in terms of the functional groups present in the molecule (Fig.~\ref{fig:turing_test}g).


%


In summary, we find that: (i) even subjects with university-level training in organic chemistry fail to correctly identify real molecules in close to four out of ten attempts; (ii) for some particular molecules (acyclic and predominantly aliphatic), performance is actually compatible with random guessing; and (iii) there are no particular classes of molecules that are systematically wrong and easy to spot, which would indicate a clear bias. These suggest that \name captures the underlying structural patterns and chemical relationships of real molecules with high fidelity, while still exploring novel regions of chemical space.

\section*{Discussion}

We have introduced \name, a collaborative constrained discrete diffusion model capable of generating novel molecules that are guaranteed to be chemically valid, and with physicochemical properties distributed very similarly to known real molecules. By enforcing valence constraints through the double edge swapping mechanism and incorporating a collaborative time model, we have addressed key challenges that have limited previous molecular generative approaches. The result is a model that achieves perfect chemical validity and generates diverse and realistic molecular structures, as verified through comprehensive benchmarks and expert evaluation.

Indeed, where previous models like JTVAE and DiGress attempt to learn chemical rules into their parameters, \name imposes constraints directly into the generative process itself. This eliminates the possibility of generating invalid molecules by design, not by training. This architectural choice transforms how molecular generation is performed, shifting the responsibility for chemical validity from the model parameters to the diffusion process itself.
Because validity is guaranteed by construction, our design philosophy  enables us to create a remarkably efficient model, which in its simplest form has only 534K  parameters, compared to DiGress' 4.6 million and JTVAE's 5.3 million parameters.
By embedding chemical knowledge into the generative process rather than the parameters, we give our model the freedom to focus entirely on capturing the subtle structural patterns of real molecules. The advantage of this approach becomes apparent when comparing the distributions of the physicochemical properties of the generated molecules, for which even our BASE model achieves a higher KL divergence score than DiGress and JTVAE, and  our FPS model outperforms DiGress on 64\% of properties and JTVAE on 92\% of properties. 

The collaborative mechanism between our diffusion and time models also proved to be a critical innovation. Initial experiments using just the diffusion model with the actual time step as input showed limited effectiveness because the denoising process progresses at different rates for different molecules. The time model resolves this by learning to predict how far a molecular graph is from completion, providing a more informed measure of progress. Unlike previous approaches that use predetermined schedules, our model adapts its predictions to the actual state of the molecule, resulting in a more precise generation of molecules. 

Finally, to show \name's efficient exploration of the chemical space, we generated a database of 8.2 million synthetic molecules, with 7.1\% redundancy and 98.5\% novelty. This demonstrates that our constrained approach does not limit molecular diversity. Traditional approaches waste computational resources generating and then filtering out invalid structures, while \name focuses exclusively on the valid regions of chemical space from the outset. We consider that this database will be a valuable resource for the community, allowing for 
the exploration of new chemical spaces and addressing new and existing challenges in molecular engineering.

Some aspects of \name may need to be modified in certain situations. First, its design implies that molecular formula does not change during the noising or denoising diffusion processes---in situations where the formula is part of the design space, this may be limiting. In such situations, however, one can always design a ``seeding algorithm'' to generate valid molecular formulas (a task that is simpler than the generation of whole molecules), and then feed those to \name.
Another aspect that may be limiting is computational complexity. Because each step involves four bonds, the computational complexity of \name's diffusion steps is \(O(n^4)\), where $n$ is the number of atoms in the molecule. Since not all quadruplets of bonds are suitable for our double swap move, changes in implementation and architecture may reduce this worst-case complexity considerably in practice. However, even in its current implementation, our model is able to generate molecules with up to 70 atoms in mid-range GPUs, which is of the same order as other generative models.


Future work to better understand the predictions of the model and how it generates molecules would be useful to assess how it explores the chemical space \cite{amaral24}. Besides the model itself, our synthetic molecule database could be used to explore chemical space and try to discover functionalities for the molecules generated by the model \cite{ren2021}. Additionally, extending \name for conditional molecule generation based on desired properties, integrating it with reinforcement learning approaches \cite{schulman2017}, and applying it to reconstruct molecular structures from incomplete data represent exciting opportunities \cite{bohde2025}. These directions would build upon the foundation we have established here---a model that guarantees valid molecule generation with property distributions closely matching real molecules.


\section*{Methods}

\subsection*{Molecular data processing}

We use a curated dataset of 2.25 million molecules derived from several established molecular databases including PubChem, ChEMBL, ZINC, and NIST. The process of curating and processing our molecular database involved multiple steps to ensure quality and consistency.

Initially, all molecules were represented in SMILES format. Using RDKit, we canonicalized these SMILES strings based on their InChI keys to establish a standardized representation. During this process, we eliminated duplicates and molecules that could not be properly converted. We deliberately chose to work with a reduced but stable set of molecules that are consistently represented, rather than incorporating a larger quantity of molecules from different datasets with inconsistent representations. This approach ensures that our model learns from a clean, uniform dataset rather than having to accommodate representation inconsistencies that might exist across different molecular collections.

After canonicalization, the SMILES strings were converted to molecular graphs composed of nodes (atoms) and edges (bonds). For each molecule, we extract the explicit heavy atoms from the RDKit molecule object. Additionally, implicit hydrogen atoms are derived and represented as explicit nodes in our graph. This approach treats all atoms, including hydrogens, as first-class entities in the molecular graph.

To manage computational complexity, we restricted our dataset to molecules containing between 5 and 70 atoms. This upper limit was established due to architectural constraints in our diffusion model. Specifically, since our model needs to calculate the probabilities for every possible double edge swap, the complexity scales as \(O(n^4)\) with the number of atoms \(n\). For a molecule with 70 atoms, this already represents a substantial computational space; including larger molecules would increase both memory requirements and processing time. After applying this size filter, approximately 1.67 million molecules remained in our dataset.



\subsection*{Molecular Graph Diffusion}

As introduced in the main text, our approach uses a discrete diffusion process based on double edge swapping (DES) that preserves atomic valence constraints throughout the diffusion trajectory, ensuring chemical validity. Here we describe the mathematical formulation of the noising process and the denoising process. The details about how the collaborative interaction between our diffusion and time models enables efficient generation of valid molecules are provided later, in section ``Sampling of new molecules.'' 

\paragraph{Noising process}

Our diffusion process is built upon a valence-constrained mechanism that ensures all molecular graphs throughout the diffusion trajectory maintain chemical validity. The core of this mechanism is the DES operation, in which we: (i) randomly select two edges $e_1 = (i,j)$ and $e_2 = (k,l)$ in the molecular graph $G_t$; (ii) remove these edges; (iii) create two new edges $e_3 = (i,k)$ and $e_4 = (j,l)$ by cross-connecting the atoms of the original edges. This process ensures that each atom maintains its original valence because each atom loses one bond and gains another. By iterating this process, the molecular graph diffuses toward a Molloy-Reed distribution~\cite{molloy95,maslov02,milo04}, which is the maximum-entropy distribution over graphs with a fixed degree sequence.

Mathematically, the DES operation can be described as a transformation $T: G_{t-1} \rightarrow G_{t}$. 
Since each DES operation affects four bonds (two bonds are removed and two new bonds are created), the number of DES operations needed to completely randomize a molecule is approximately 25\% of the total number of bonds. 
%
Unlike existing discrete diffusion models, which use a constant transition matrix, our diffusion approach is a Markov process where the transition probability from a molecular graph $G_{t-1}$ to a noisier graph $G_t$ depends on $G_{t-1}$. Therefore, there is no general closed-form expression for the $t$-step transition matrix.
The multidimensional transition matrix $Q_t$ has elements $[Q_t]_{ijkl}$, which represent the probability that an edge $(i,j)$ and an edge $(k,l)$ are removed (note that there might be other edges remaining between $(i,j)$ and/or $(k,l)$ if the original multiplicity of those edges was larger than one, that is, if the bonds were double or triple) and edges $(i,k)$ and $(j,l)$ are created (note that edges $(i, k)$ and/or $(j, l)$ may already exist, in which case we simply increase the multiplicity of such edges), and are given by
\[
    [Q_t]_{ijkl} = \frac{F_t(i, j, k, l)}{\sum_{i', j', k', l'} F_t(i', j', k', l')}
\]
Where $F_t(i, j, k, l)$ is function that determines the feasibility of removing edges $(i,j)$ and $(k,l)$ and creating valid edges $(i,k)$ and $(j,l)$ given molecular graph $G_t$. 
%
%
This function is defined as
\[
F_t(i, j, k, l) = \begin{cases} 1 & \text{if removing $(i,j)$ and $(k,l)$ and creating $(i,k)$ and $(j,l)$ results in a valid molecular graph} \\ 0 & \text{otherwise} \end{cases}
\]
For $F_t(i, j, k, l)$ to equal 1, the following conditions must be met:
\begin{enumerate}
    \item All four nodes must be distinct, $i \neq j \neq k \neq l$.
    \item Edges $(i, j)$ and $(k, l)$ must exist in $G_t$.
    \item The new edges must give rise to, at most, triple bonds.
    \item The connectivity of the resulting graph $G_{t+1}$ must be maintained.
\end{enumerate}




\paragraph{Denoising process}

The denoising process in our molecular graph diffusion is mathematically formalized as an optimization process that seeks to reverse the structural modifications introduced during the noising process. More precisely, let $G_t$ be the molecular graph at time $t$ during the diffusion process, and $G_0$ be the original molecular graph. The objective of the denoising process is to find a sequence of transformations $T^{-1}: G_t \rightarrow G_{t-1}$ that reverse the structural modifications and recover a chemically valid molecular graph with properties similar to those of real molecules.
This process is implemented through our constrained collaborative mechanism, which employs two specialized models that work in tandem---a {\it diffusion model} and a {\it time model}.

Each denoising step takes as input the molecular graph $G_t$ and the normalized time step $t$, and selects a suitable DES. To do this, the diffusion model learns three probability distributions: (i) the probability $[Q_t^{-1} (\theta, \theta_f, \theta_b)]_{ijkl} = {\rm Prob}_{\theta, \theta_f, \theta_b}(\text{select } (i,j) \; \& \; (k,l) | G_t, t)$ of selecting $(i,j)$ and $(k,l)$ for the next denoising DES; (ii) the probability $[P_t^{\rm form} (\theta, \theta_{f})]_{ij} = {\rm Prob}_{\theta, \theta_f}((i, j) \; \text{exists} \; | G_t, t)$ of forming an edge $(i, j)$; (iii) the probability $[P_t^{\rm break} (\theta, \theta_{b})]_{ij} = {\rm Prob}_{\theta, \theta_b}((i, j) \; \text{does not exist} \; | G_t, t)$ of breaking an edge $(i, j)$. 
%


Some parameters $\theta$ of these distributions are shared among all models, whereas others ($\theta_f, \theta_b$) are specific to different distributions. These parameters are learned so as to minimize three corresponding binary cross-entropy loss functions. For DES prediction, we minimize
\begin{equation}
    \mathcal{L}_{\text{BCE-DES}} = -\frac{1}{N_q} \sum_{(i,j,k,l)} \left[ y_{ijkl}^{t-1} \log [Q_t^{-1} (\theta, \theta_f, \theta_b)]_{ijkl} + (1 - y_{ijkl}^{t-1}) \log \left( 1 - [Q_t^{-1} (\theta, \theta_f, \theta_b)]_{ijkl} \right) \right]
    \label{eq:lossq}
\end{equation}
where $N_q$ is the number of feasible quadruplets $(i, j, k, l)$, $y_{ijkl}^{t-1}$ is the binary label indicating whether DES $(i,j)$ and $(k,l)$ is the one that actually led from $G_{t-1}$ to $G_t$ during the (forward) noising process.

For bond formation prediction we minimize
\begin{equation}
    \mathcal{L}_{\text{BCE-form}} = -\frac{1}{N_P} \sum_{(i,j)} \left[ y_{ij}^{f0} \log [P_t^{\rm form} (\theta, \theta_{f})]_{ij} + (1 - y_{ij}^{f0}) \log \left( 1 - [P_t^{\rm form} (\theta, \theta_{f})]_{ij} \right) \right]
    \label{eq:lossf}
\end{equation}
where $N_P$ is the number of pairs of nodes in the molecular graph, and $y_{ij}^{f0}$ indicates whether edge $(i,j)$ should be formed with respect to the original molecule (time $t=0$).

Finally, for bond breakage prediction we minimize
\begin{equation}
    \mathcal{L}_{\text{BCE-break}} = -\frac{1}{N_E} \sum_{(i,j)} \left[ y_{ij}^{b0} \log [P_t^{\rm break} (\theta, \theta_{b})]_{ij} + (1- y_{ij}^{b0}) \log \left( 1 - [P_t^{\rm break} (\theta, \theta_{b})]_{ij} \right) \right]
    \label{eq:lossb}
\end{equation}
where $N_E$ is the number of pairs of edges in the molecular graph, and $y_{ij}^{b0}$ indicates whether edge $(i,j)$ should be broken with respect to the original molecule (time $t=0$).

The denoising process also requires a time model, which learns to predict the normalized time step of the diffusion process. This model takes as input the molecular graph $G_t$, node features $X$, edge features $E$, and the graph features $g$, and estimates how far the current molecular graph is from the original molecule, providing a normalized time value between 0 and 1.
The time model is trained using mean squared error loss
\begin{equation}
    \mathcal{L}_{\text{MSE}} = \left( t_{\text{pred}} - t_{\text{real}} \right)^2
    \label{eq:losst}
\end{equation}

Below, in section ``Sampling of new molecules'', we describe how these models are used in the actual process of sampling, that is, of generating new molecules.




\subsection*{Diffusion model and time model architectures}

\begin{figure}[t!]
  \centering
  \includegraphics[width=1\textwidth]{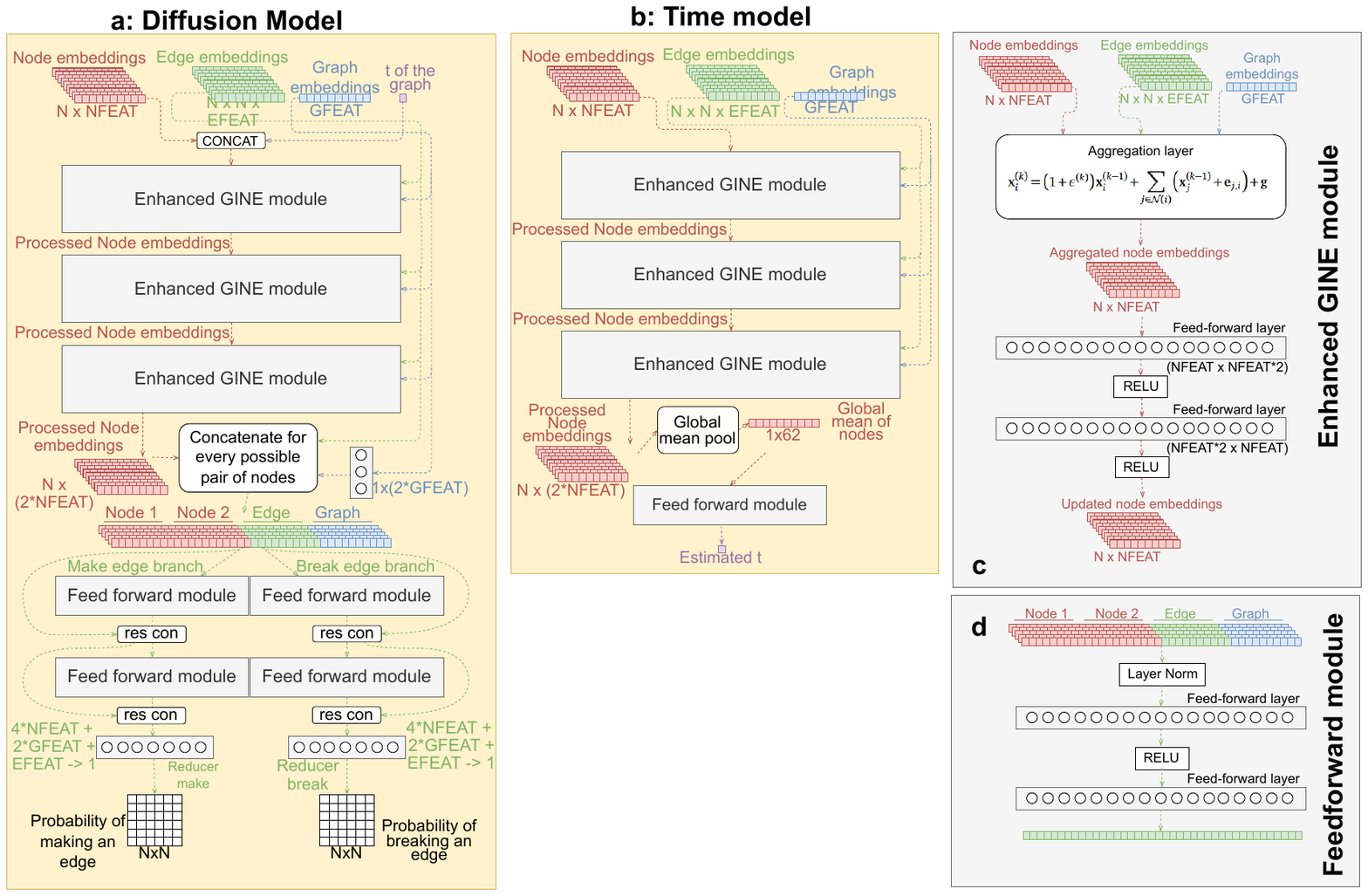} 
  \caption{{\bf Architecture of \name components.} {\bf a}, The diffusion model processes the molecular graph through a sequence of EnhancedGINE layers, the embedding of pairs of nodes are concatenated with edge properties and processed through two feedforward modules to predict the probability of bond formation and bond breakage for each possible double edge swapping operation. {\bf b}, The time model estimates the diffusion timestep $t$ of the current molecular graph using processed node embeddings obtained after applying the EnhancedGINE module to the features of the molecular graph. {\bf c}, The message passing component of both models, the EnhancedGINE module. {\bf d}, The prediction component of the diffusion model.}
  \label{fig:model_architecture}
\end{figure}

Our collaborative constrained diffusion approach is implemented through two separate neural network architectures that process molecular graph features and work together during inference---the diffusion model and the time model. These architectures are designed to efficiently handle the graph-structured data while keeping the parameter count low. The valence constraints are enforced by the DES mechanism rather than by the neural architectures, which just make the predictions for pairs of edges to choose during the denoising process. Figure~\ref{fig:model_architecture} provides a comprehensive overview of our model architectures for the BASE \name model.

\paragraph{Diffusion model architecture}

The diffusion model employs a graph neural network (GNN) architecture consisting of two main components: (i) a message-passing component that processes node),  edge and graph features and the diffusion time; and (ii) a prediction component that estimates edge probabilities from processed features (Fig.~\ref{fig:model_architecture}a).

The most important component of the diffusion model is the message-passing component (Fig.~\ref{fig:model_architecture}c), which uses a sequence of three enhanced graph isomorphism network \cite{hu2020} with edge features (EnhancedGINE) layers. These layers extend the standard GINE layers by incorporating global graph features directly into the message-passing mechanism. Each EnhancedGINE layer transforms node features into 124-dimensional embeddings, with the final layer outputs capturing comprehensive atomic environments. The architecture can be summarized as follows:
\begin{enumerate}
    \item \textbf{Initial feature processing}: Node features $X$, edge features $E$, graph features $g$ and the diffusion time $t$ are processed by the first EnhancedGINE layer.
    \item \textbf{Hidden representations}: The output embeddings are passed through a non-linear activation function followed by a feedforward layer. This is repeated for the second and third EnhancedGINE layers, with residual connections between layers to preserve information flow.
    \item \textbf{Node embedding aggregation}: After the EnhancedGINE layers, we obtain embeddings for each node that capture its local and global context within the molecular graph.
    \item \textbf{Edge probability}: For each pair of nodes, the corresponding node embeddings are concatenated to each other and to edge and graph features, and processed through two different feedforward networks, each with 256 hidden units (Fig.~\ref{fig:model_architecture}d), to predict: (i) the probability of bond formation between each pair of atoms; (ii) the probability of bond breakage for existing bonds.
    \item \textbf{Double edge swap probability}: These individual bond probabilities are combined to compute the probability of each possible double edge swapping operation.
\end{enumerate}

The probabilities for the DES are computed outside of the model by combining the probabilities of bond formation and bond breakage for each possible double edge swapping operation. This architecture efficiently processes molecular graphs with approximately 471K parameters for the BASE model. The neural network makes no assumptions about chemical validity---it focuses solely on learning structural patterns from the training data, while the valence constraints are handled externally by the diffusion process.

\paragraph{Time model architecture}

The time model shares a similar GNN backbone with the diffusion model but serves a different purpose. It estimates the progress of the diffusion process by predicting how far the current molecular graph is from a valid molecule. Its architecture consists of (Fig.~\ref{fig:model_architecture}b):
\begin{enumerate}
    \item \textbf{EnhancedGINE layers}: Three layers that process node, edge, and graph features similar to the diffusion model, maintaining the same 124-dimensional embeddings.
    \item \textbf{Graph-level embedding}: Node embeddings are aggregated using mean pooling to obtain a fixed-dimensional representation (124-dimensional) of the entire molecular graph.
    \item \textbf{Time prediction}: The graph embedding is processed through a simple feedforward network (64 hidden units) that outputs a scalar value between 0 and 1, representing the normalized diffusion time.
\end{enumerate}

With approximately 63K parameters, the time model is significantly smaller than the diffusion model while still providing crucial guidance during the denoising process.

\paragraph{FPS model variant}

\begin{figure}[t!]
  \centering
  \includegraphics[width=1\textwidth]{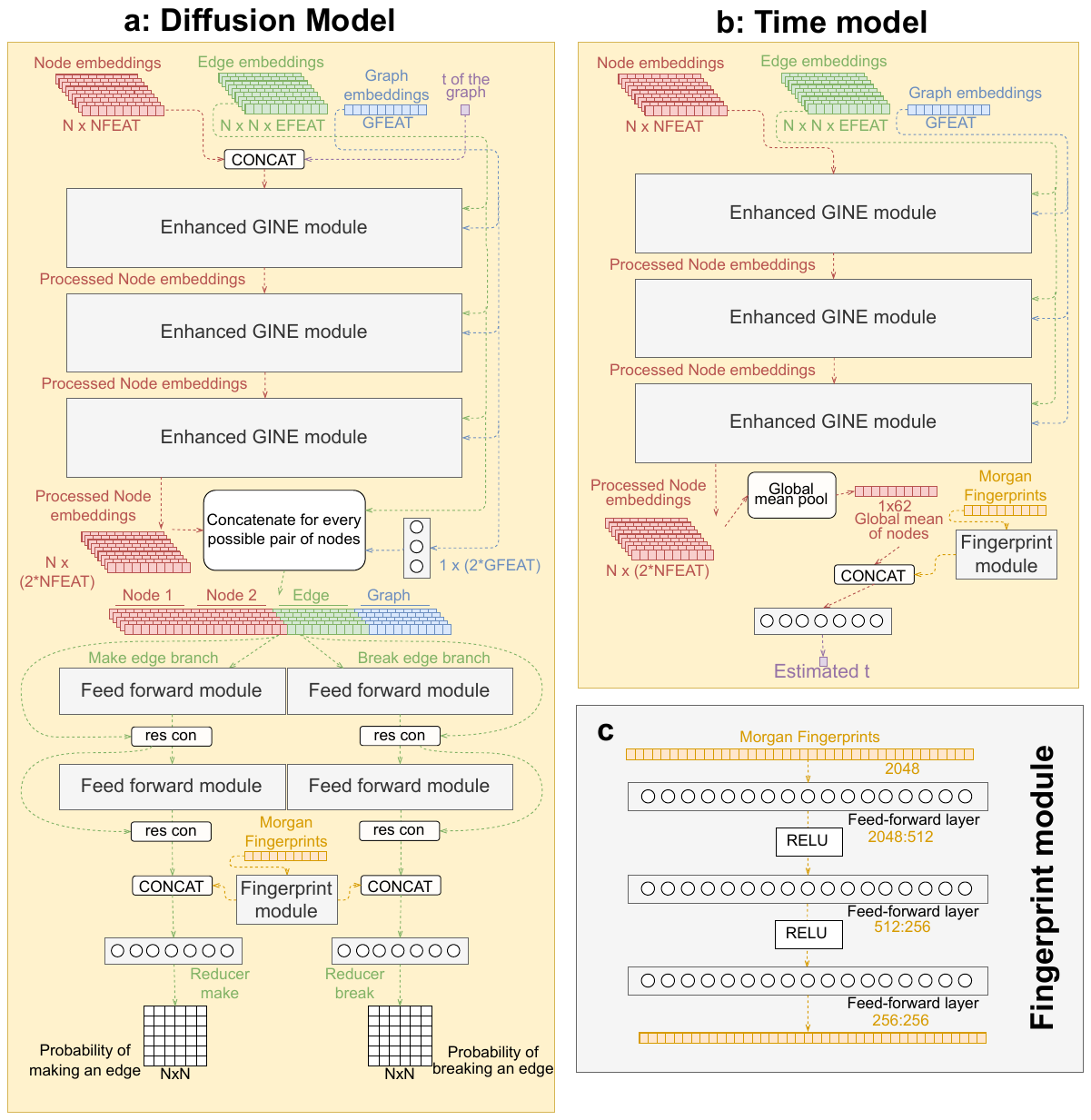} 
  \caption{{\bf Architecture of \name FPS (fingerprint-enhanced) components.} {\bf a},  The diffusion model processes the molecular graph through a sequence of EnhancedGINE layers. Then, the embedding of pairs of nodes are concatenated with edge and graph properties and processed through two feedforward modules, after which the fingerprint processed by the fingerprint module is concatenated. The resulting vector is processed through a feedforward network to predict the probability of bond formation and bond breakage for each possible double edge swapping operation {\bf b}, The time model estimates the diffusion timestep $t$ of the current molecular graph using processed node embeddings concatenated with the fingerprint processed by the fingerprint module. {\bf c}, The fingerprint module of both models.}
  \label{fig:model_architecture_fps}
\end{figure}

The fingerprint-enhanced (FPS) variant of \name extends the BASE models by incorporating molecular fingerprints that capture substructural information. This architecture, depicted in Fig.~\ref{fig:model_architecture_fps}, processes the 2048-dimensional Morgan fingerprints (ECFP3) through:
\begin{enumerate}
    \item \textbf{Fingerprint processing}: The binary fingerprint vector is passed through a dedicated feedforward neural network (with layers of 1024, 512, and 256 units) that reduces its dimensionality while preserving substructure information.
    \item \textbf{Feature integration}: The processed fingerprint features (256-dimensional) are concatenated with the graph embeddings prior to the final prediction layers of the diffusion and time models.
    \item \textbf{Enhanced prediction}: The combined representations enable the models to make predictions informed by both graph structure and specific molecular substructures.
\end{enumerate}

This enhancement increases the parameter count to approximately 3.1M for the diffusion model and 1.3M for the time model, but improves performance by incorporating explicit substructural information.

\subsection*{Molecule featurization}

Molecule featurization transforms molecular information into structured numerical data that can be processed effectively by our models. By extracting relevant features at the node (atom), edge (bond), and graph (molecule) levels, we allow \name models to accurately capture the molecular properties that characterize valid chemical structures. These features provide a comprehensive representation of molecular characteristics at multiple levels of granularity. While additional topological and structural features could be extracted, our experiments suggest that this feature set strikes a balance between model performance and computational efficiency. In Table~\ref{tab:features}, we summarize the molecular features used by \name, which we describe in more detail next.
\begin{table}[t!]
\centering
\begin{tabular}{lccp{8cm}}
\hline
\textbf{Feature} & \textbf{Type} & \textbf{Dimensions} & \textbf{Description} \\ \hline
\\
\multicolumn{4}{l}{\textit{Node (atom) features}} \\ \hline
Element type & Node & 15 & One-hot encoding of atom elements (C, H, N, O, F, etc.) \\ \hline
Cycle participation & Node & 13 & Binary indicators for presence in cycles of size 3-14 and 15+ \\ \hline
Heavy atom neighbors & Node & 1 & Number of non-hydrogen connected atoms \\ \hline
Hydrogen neighbors & Node & 1 & Number of hydrogen atoms connected \\ \hline
Bridge count & Node & 1 & Number of bridges the atom participates in \\ \hline
\\
\multicolumn{4}{l}{\textit{Edge (bond) features}} \\ \hline
Bond multiplicity & Edge & 4 & One-hot encoding of bond type (single, double, triple, aromatic) \\ \hline
Cycle participation & Edge & 13 & Binary indicators for presence in cycles of size 3-14 and 15+ \\ \hline
Bridge status & Edge & 1 & Whether the bond is a bridge (1) or not (0) \\ \hline
Path count & Edge & 1 & Number of distinct paths between bonded atoms \\ \hline
2D distance & Edge & 1 & Spatial distance between atoms in 2D coordinates \\ \hline
\\
\multicolumn{4}{l}{\textit{Graph (molecule) features}} \\ \hline
Cycle counts & Graph & 13 & Number of cycles of size 3-14 and 15+ \\ \hline
Planarity & Graph & 1 & Measure of molecular graph planarity \\ \hline
Connected components & Graph & 1 & Number of distinct connected subgraphs \\ \hline
Bridge fraction & Graph & 1 & Fraction of edges that are bridges \\ \hline
Simplified bridge fraction & Graph & 1 & Fraction of edges that are simplified bridges \\ \hline
Bond Type Distribution & Graph & 4 & Proportion of each bond type in the molecule \\ \hline
\\
\multicolumn{4}{l}{\textit{Diffusion process features}} \\ \hline
Normalized diffusion time & Process & 1 & Normalized time step in the diffusion process (0-1) \\ \hline
\\
\multicolumn{4}{l}{\textit{FPS model additional features}} \\ \hline
Morgan fingerprints (ECFP3) & Fingerprint & 2048 & Binary representation of molecular substructures \\ \hline
\end{tabular}
\caption{{\bf Features used for molecular representation in \name models}. Features are divided in 5 groups: node (atom) features, edge (bond) features, graph (molecule) features, diffusion process features (used only in the diffusion model), and FPS features (used only in \name FPS models). }
\label{tab:features}
\end{table}

\paragraph{Node-level features}

At the node level, we extract features $X$ that encode both chemical and structural properties of each atom: (i) one-hot encoded representation of the atom's element (15 dimensions covering the most common elements in organic chemistry, namely, boron, nitrogen, carbon, oxygen, fluorine, phosphorus, sulfur, chlorine, bromine, iodine, calcium, potassium, sodium, magnesium, and hydrogen); (ii) binary indicators showing presence in cycles of sizes 3 to 14, and larger than 14; (iii) number of non-hydrogen neighboring atoms; (iv) number of bridges the atom participates in.
These node features enable the model to understand the chemical environment around each atom when predicting valid double edge swapping operations.

\paragraph{Edge-level features}

For edges, we extract features $E$ that describe bond properties and their structural role in the molecular graph: (i) one-hot encoding of bond multiplicity (single, double or triple); (ii) binary features indicating participation in cycles of sizes 3 to 14 and more than 14; (iii) binary indicator of whether the bond is a bridge; (iv) number of distinct paths between the two atoms of the bond; (v) 2D distance between the atoms in the molecular graph.
These edge features allow the diffusion model to identify bonds that can be validly swapped while maintaining chemical constraints.

\paragraph {Graph-level features}
At the whole-graph level, we capture global structural properties $g$: (i) number of cycles between sizes 3 to 14 and above 14; (ii) a measure of whether the molecular graph is planar; (iii) number of connected components; (iv) fraction of edges that are bridges and simplified bridges; (v) proportion of each bond multiplicity in the molecule.
Additionally, the diffusion time step is stored as a normalized feature between 0 and 1, providing temporal context during the diffusion process.

\paragraph{Molecular fingerprints in FPS models}
For our enhanced \name FPS model, we incorporate Morgan fingerprints \cite{Morgan1965} with a dimensionality of 2048. These fingerprints capture the presence of specific substructures within the molecule, providing a rich representation of molecular motifs that may not be explicitly captured by the other features. The inclusion of these fingerprints allows the FPS \name model to identify patterns of substructural arrangements that correlate with valid chemical transformations, enhancing its ability to predict realistic double edge swapping operations.

\paragraph{Feature normalization}
All extracted features are normalized to ensure balanced contribution to the model's learning process. For categorical features such as element type and bond multiplicity, we use one-hot encoding. Count-based features are normalized to appropriate ranges, while binary features remain as 0/1 indicators. 


\subsection*{Model training}

Our training dataset consisted of 2.25 million molecules derived from established molecular databases as described in the `Molecular processing' section. For computational efficiency, we processed the dataset in slices of approximately 100,000 molecules, with 80\% allocated for training and 20\% for validation. Molecules were filtered based on our established criteria (5-70 atoms, valid chemical elements, and a single connected component).

The training process leveraged an implicit form of data augmentation arising from our diffusion methodology. While the dataset contained a finite number of molecules, our constrained diffusion process generated a different random intermediate graph for each molecule during each training iteration through the application of random double edge swapping operations. This approach effectively expanded the training distribution, enhanced generalization capabilities, and prevented overfitting to specific diffusion trajectories.

\paragraph{Diffusion Model Training}

We trained the diffusion model using a step-wise approach with batch size 12. 
For each step in the diffusion trajectory, we computed features as detailed in the `Feature Extraction' section. We trained the model by processing pairs of consecutive diffusion steps, which yielded better performance compared to batch-wise or molecule-wise training.

The training used three weighted binary cross-entropy loss components, as described above.
%
Each loss was weighed and masked and applied to the model normalized by the number of diffusion steps processed in each backward pass to ensure stable training regardless of molecular size or diffusion trajectory length.

\paragraph{Time Model Training}

We trained the time model using a similar data processing approach to the diffusion model, with identical batch sizes and dataset slicing. Unlike the diffusion model, which predicts edge swapping operations, the time model was trained to predict the normalized diffusion time (ranging from 0 to 1) corresponding to each graph state in the diffusion trajectory.

We employed a simple mean squared error (MSE) loss between the predicted normalized time and the actual time step of the diffusion process. This model provides crucial guidance during the denoising process by informing the diffusion model of the actual progress of denoising, as illustrated in Fig.~\ref{fig:model}.

\paragraph{Optimization and computational implementation}

The BASE variant of the diffusion and time models were  trained for three epochs with an initial learning rate of $10^{-4}$. The FPS models were initialized with 1 epoch pre-trained BASE weights and fine-tuned for 2 more epochs using differential learning rates---$10^{-5}$ for pre-trained parameters and $10^{-4}$ for the new fingerprint-related parameters, preserving learned knowledge while allowing fingerprint-specific parameters to adapt more rapidly.

To manage the computational load, we implemented several efficiency measures, including distributed dataset processing with checkpoints saved every 1,000 batches, state preservation across slice boundaries, and parallel feature computation using a process pool executor with 24 workers. The complete training process for all model variants required approximately 60 days on a single NVIDIA RTX 4090 GPU. 

\subsection*{Sampling of new molecules}

For the sampling of new molecules, we employ our collaborative constrained diffusion model. During the sampling process, both the diffusion and time models work together to generate the final molecule. Initially, a molecular formula is selected directly from the database. This molecular formula determines the atoms (nodes) and their valences (degree constraints) for the generation process. A random molecular graph that satisfies these constraints is then constructed to obtain \( G_T \) using the noising process as described above.

During the denoising process, a cycle is repeated where the diffusion and time models are applied iteratively to generate the final molecule. At step $t$, the molecular graph features are extracted, and the diffusion model is applied to obtain the probabilities of double edge swapping operations, bond breakage, and bond formation. Based on these probabilities, a random selection is made from the possible double edge swapping operations with a probability greater than a threshold (initially set at 95\%). If the threshold prevents any operation from being selected, the threshold is reduced by 5\%. 

Once a double edge swapping operation is selected and applied, the resulting graph is verified to ensure that it has not been previously encountered in the sampling process and that it remains fully connected---a critical requirement for valid molecules. If these conditions are met, the time model is applied to predict the normalized diffusion time \( t_{\text{pred}} \) of the current graph state. This cycle is repeated for a predetermined number of diffusion steps, calculated based on the number of bonds in the molecular formula.

Finally, from the trajectory of molecular graphs generated from \( G_T \) along with their diffusion time predictions, the output molecule is selected as the graph with the normalized diffusion time \( t_{\text{pred}} \) closest to 0. This selection mechanism, which prioritizes graphs that the time model identifies as being closest to valid molecules rather than simply taking the final graph in the sequence, is a key advantage of our collaborative approach. The result is a chemically valid molecule with the specified molecular formula, generated through a controlled and constrained diffusion process, and validated by the collaboration between our diffusion and time models.


\subsection*{Development of Turing-test web and analysis of results}

To evaluate the plausibility of our synthetic molecules, we developed a web-based molecular Turing-like test platform. This platform was designed to present human experts with pairs of molecules and challenge them to distinguish between real molecules from our database and novel molecules generated by \name. 
%
%
The application was hosted on our laboratory server and made accessible through the URL \url{http://cocograph.seeslab.net}.
%

The molecule pairs were drawn from our database of 8.2 million generated molecules and their corresponding original molecules used as seed structures. For each testing round, the application randomly selected molecule pairs. Molecules were rendered as 2D structural diagrams using RDKit's drawing utilities to ensure clarity and consistency in presentation.

\paragraph{Test protocol and participant recruitment}

Participants were presented with an information page explaining the nature of the experiment without revealing specific details about the generative model or selection criteria that might bias their choices. After consenting to participate and providing their expertise level (high school, undergraduate or postgraduate training in organic chemistry), each participant proceeded through 20 rounds of molecule comparison. In each round:

\begin{enumerate}
    \item Two molecular structures with identical molecular formulas were displayed side by side.
    \item The order of real and generated molecules was randomized for each pair.
    \item Participants selected which molecule they believed was from an established chemical database.
    \item No time limit was imposed, allowing participants to make thorough evaluations.
    \item No feedback was provided until completion of all 20 rounds to prevent learning effects.
\end{enumerate}

We recruited participants primarily from the University Rovira i Virgili, targeting departments of chemical engineering, mechanical engineering, and the faculties of chemistry and biochemistry. The recruitment process involved email invitations to departmental mailing lists and direct outreach to research groups. The test was accessible for a two-week period, allowing participants to complete it at their convenience.

\paragraph{Data analysis methodology}

For each participant, we recorded their level of expertise and their selections for each molecule pair. 
The primary metric calculated was accuracy---the percentage of molecule pairs for which participants correctly identified the real molecule. We analyzed this metric globally and stratified by: (i) level of expertise; (ii) molecular size; and (iii) chemical properties, including functional groups, aromaticity, and other structural features calculated using RDKit.
Confidence intervals (95\%) for accuracy metrics were calculated using bootstrap with 1,000 resampling iterations to ensure robust statistical inference.


\bibliography{references}

\section*{Acknowledgements}
This research was supported by project PID2022-142600NB-I00 (M.SP. and R.G.) from MCIN/ AEI/ 10.13039/ 501100011033, and by the Government of Catalonia (2021SGR-633) (M.SP. and R.G.). The authors thankfully acknowledge the computer resources at MareNostrum5, and the technical support provided by BSC (RES-IM-2025-1-0021).

\section*{Author contributions statement}
All authors designed research and studied current state of the art in molecular generation models.
M.R-B. wrote code for processing the data, training models, and sampling new molecules, as well as the web application for the Turing-like test.
All authors analyzed results, discussed evolution of the project and wrote the paper.

\subsection*{Data availability}
Data used for training was obtained from several databases of molecular compounds, then filtererd and processed. This data and generated database of new synthetic molecules is available at: \url{https://github.com/manurubo/CoCoGraph/tree/main}.

\subsection*{Code availability}

Our code for training our constrained collaborative model, sampling new molecules and evaluate the results is available at: \url{https://github.com/manurubo/CoCoGraph/tree/main}.
The weights of the models are available at the same repository. 

\section*{Competing interests}
The authors declare no competing interests.

\newpage
\section*{Extended Data}

\renewcommand{\thetable}{ED\arabic{table}}
\setcounter{table}{0}
\renewcommand{\thefigure}{ED\arabic{figure}}
\setcounter{figure}{0}
\begin{table}[h!]
\centering
\begin{tabular}{|l|c|c|c|}
\hline
\textbf{Property} & \textbf{CoCoGraph} & \textbf{DiGress} & \textbf{JTVAE} \\
\hline
\multicolumn{4}{|l|}{\textit{Basic Physicochemical Properties}} \\
\hline
Molecular Weight & \textbf{0.0144} & 0.0789 & 0.4858 \\
Exact Molecular Weight & \textbf{0.0133} & 0.0790 & 0.4856 \\
Heavy Atom Count & \textbf{0.0162} & 0.0812 & 0.4666 \\
Number of Valence Electrons & \textbf{0.0171} & 0.0742 & 0.4651 \\
N-H/OH Count & 0.1861 & \textbf{0.0259} & 0.0911 \\
N-O Count & \textbf{0.0175} & 0.0720 & 0.2481 \\
Fraction Csp3 & \textbf{0.0445} & 0.0499 & 0.1371 \\
Quantitative Estimate of Drug-likeness & 0.0440 & \textbf{0.0395} & 0.4617 \\
Balaban's J Index & \textbf{0.0365} & 0.0930 & 0.1889 \\
\hline
\multicolumn{4}{|l|}{\textit{Lipinski's Rule of Five Descriptors}} \\
\hline
Number of H-bond Donors & 0.1869 & \textbf{0.0247} & 0.1069 \\
Number of H-bond Acceptors & \textbf{0.0474} & 0.0819 & 0.2036 \\
Molecular LogP & \textbf{0.0304} & 0.0390 & 0.3009 \\
Number of Rotatable Bonds & \textbf{0.0590} & 0.0663 & 0.2835 \\
Topological Polar Surface Area & \textbf{0.0559} & 0.0578 & 0.2183 \\
\hline
\multicolumn{4}{|l|}{\textit{Ring and Aromaticity Descriptors}} \\
\hline
Number of Aromatic Rings & 0.0892 & \textbf{0.0202} & 0.2455 \\
Number of Aliphatic Rings & 0.1165 & \textbf{0.1152} & 0.1345 \\
Ring Count & \textbf{0.0637} & 0.1305 & 0.3074 \\
Number of Saturated Rings & \textbf{0.0541} & 0.0837 & 0.0833 \\
BertzCT & \textbf{0.0213} & 0.1051 & 0.3958 \\
\hline
\multicolumn{4}{|l|}{\textit{Electronic Descriptors}} \\
\hline
Molar Refractivity & \textbf{0.0176} & 0.0630 & 0.4642 \\
Maximum Partial Charge & 0.1502 & \textbf{0.0577} & 0.1942 \\
Minimum Partial Charge & 0.1160 & \textbf{0.0817} & 0.1579 \\
Maximum Absolute Partial Charge & 0.1098 & \textbf{0.0750} & 0.1688 \\
Minimum Absolute Partial Charge & 0.1445 & \textbf{0.0597} & 0.2170 \\
IPC & 0.0094 & 0.0073 & \textbf{0.0083} \\
EState VSA Descriptor 1 & \textbf{0.0312} & 0.0537 & 0.1546 \\
\hline
\multicolumn{4}{|l|}{\textit{Topological Descriptors (Chi Descriptors)}} \\
\hline
Chi0 Index & \textbf{0.0173} & 0.0730 & 0.4668 \\
Chi1 Index & \textbf{0.0125} & 0.0811 & 0.4627 \\
Chi2n Index & \textbf{0.0223} & 0.0876 & 0.3768 \\
Chi3n Index & \textbf{0.0337} & 0.0944 & 0.3529 \\
Chi0n Index & \textbf{0.0175} & 0.0659 & 0.4408 \\
\hline
\multicolumn{4}{|l|}{\textit{VSA (Van der Waals Surface Area) Descriptors}} \\
\hline
SlogP VSA Descriptor 1 & 0.0608 & \textbf{0.0367} & 0.1161 \\
SlogP VSA Descriptor 2 & 0.0728 & \textbf{0.0649} & 0.1581 \\
SlogP VSA Descriptor 3 & 0.0951 & \textbf{0.0448} & 0.1252 \\
SlogP VSA Descriptor 4 & 0.0814 & \textbf{0.0935} & 0.1467 \\
SlogP VSA Descriptor 5 & \textbf{0.0501} & 0.0526 & 0.1370 \\
\hline
\end{tabular}
\caption{{\bf Jensen-Shannon distances between distributions of original and generated molecules for each model across all 36 chemical properties}. Lower values indicate better distribution matching between generated and original molecules.}
\label{tab:js_supplementary}
\end{table}

\clearpage

\begin{figure}[h!]
  \centering
  \includegraphics[width=1\textwidth]{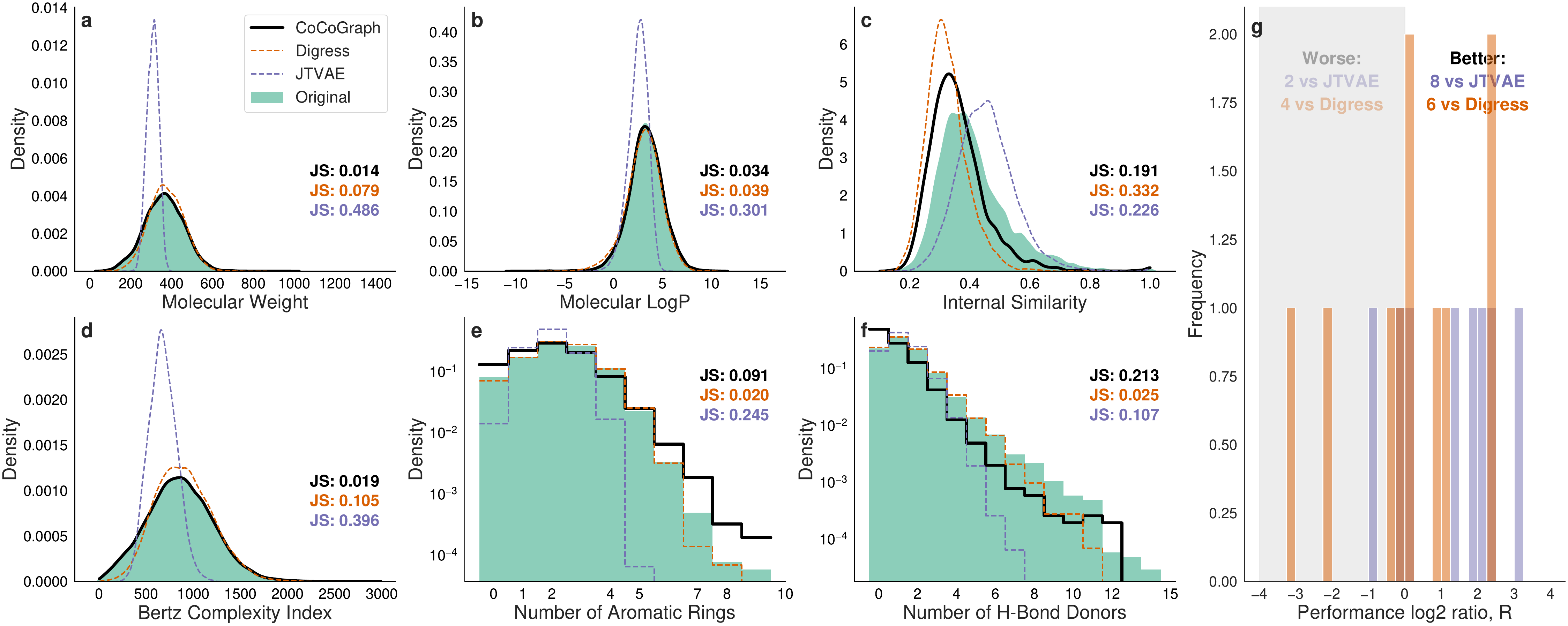} 
  \caption{{\bf Performance comparison of \name BASE on GuacaMol benchmark properties.} {\bf a-f}, Distributions of six molecular properties: {\bf a}, molecular weight; {\bf b}, molecular logP; {\bf c}, internal similarity; {\bf d}, Bertz complexity index; {\bf e}, number of aromatic rings; and {\bf f}, number of H-bond donors. For each property, the distribution of values calculated for molecules generated by \name  (black line) is compared to that of the original molecules (green distribution), and to those of molecules generated by JTVAE (purple dashed line) and DiGress (orange dashed line). Jensen-Shannon (JS) distance values between each model and the original distribution are shown. {\bf g}, Summary comparison based on the log2 ratio of JS distances between \name and comparator models for the properties in ({\bf a-f}). Positive values indicate \name outperforms the comparator model and vice versa.}
  \label{fig:combined_basemodel}
\end{figure}

\begin{figure}
  \centering
  \includegraphics[width=1\textwidth]{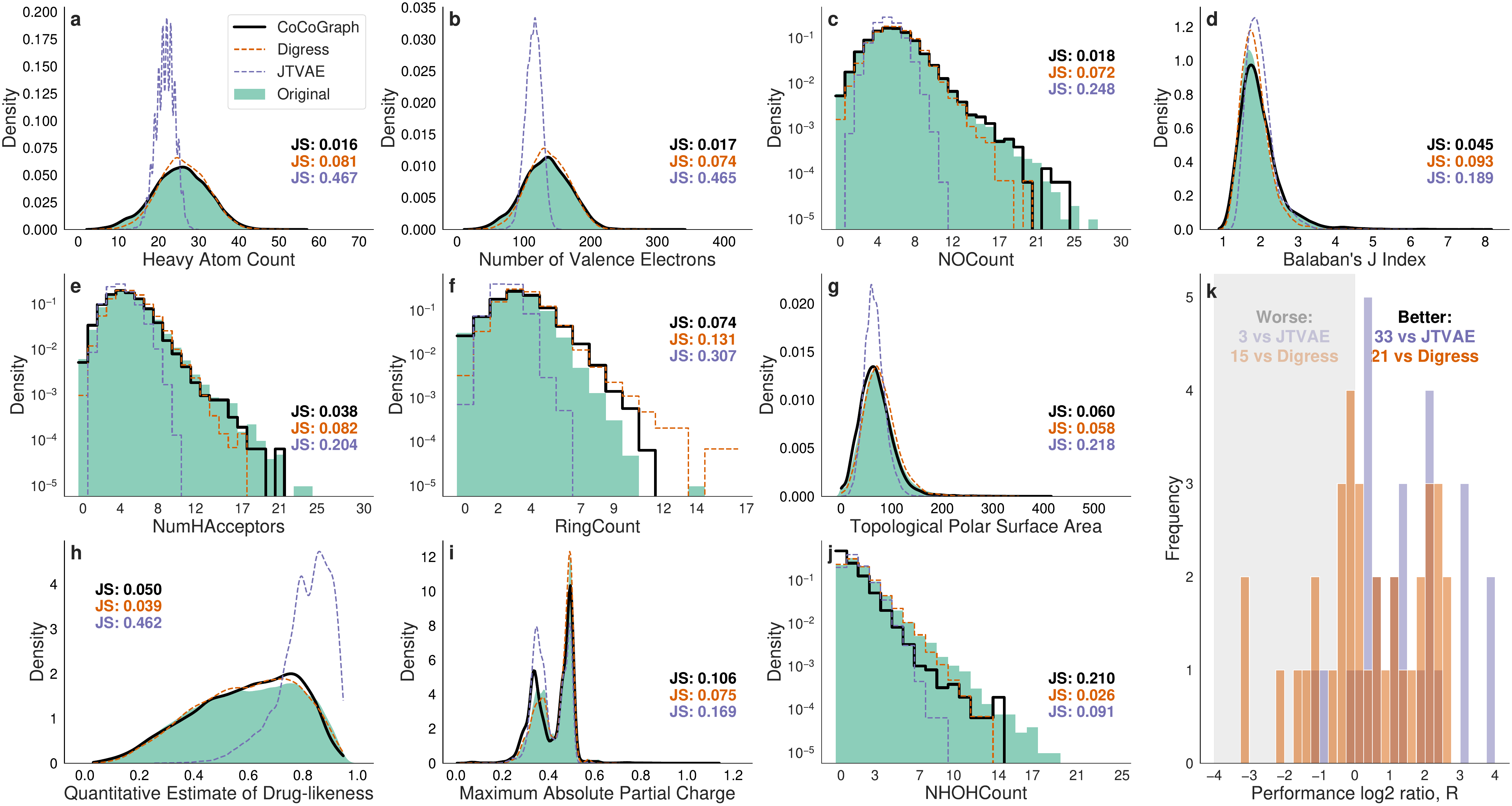} 
  \caption{{\bf Detailed performance comparison of \name BASE on a subset of 36 chemical properties.} {\bf a-j}, Distributions of ten molecular properties: {\bf a}, heavy atom count; {\bf b},  number of valence electrons; {\bf c},  NOCount; {\bf d},  Balaban’s J Index; {\bf e},  number of H acceptors; {\bf f},  ring count; {\bf g},  topological polar surface area (TPSA); {\bf h},  quantitative estimate of drug-likeness (QED); {\bf i},  maximum absolute partial charge; and {\bf j},  NHOHCount. For each property, the distributions for molecules generated by the CocoGraph FPS model (black line) is compared to that of the original molecules (green distribution) and to those of molecules generated by JTVAE (purple line) and DiGress (orange line). {\bf k},  log2 ratio of JS distances between CocoGraph FPS and the other models, where a positive value indicates that CocoGraph FPS outperforms the comparative model.}
  \label{fig:combined_basemodel_all}
\end{figure}

\begin{figure}[t!]
  \centering
  \includegraphics[width=.9\textwidth]{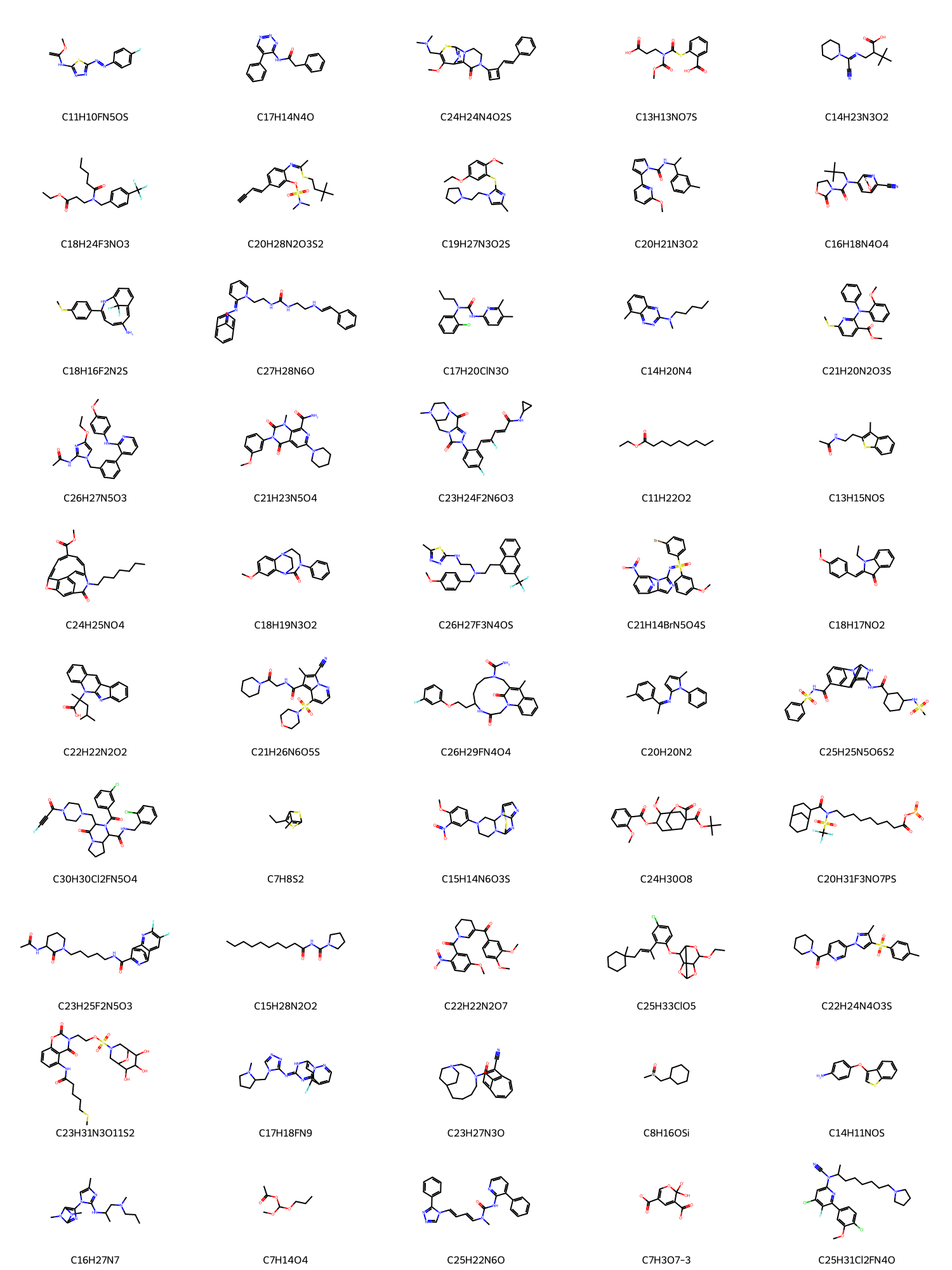} 
  \caption{{\bf 50 random molecules generated by \name FPS.} Molecules are sampled uniformly at random from our generated database.}
  \label{fig:molecule_grid}
\end{figure}

\begin{figure}
  \centering
  \includegraphics[width=1\textwidth]{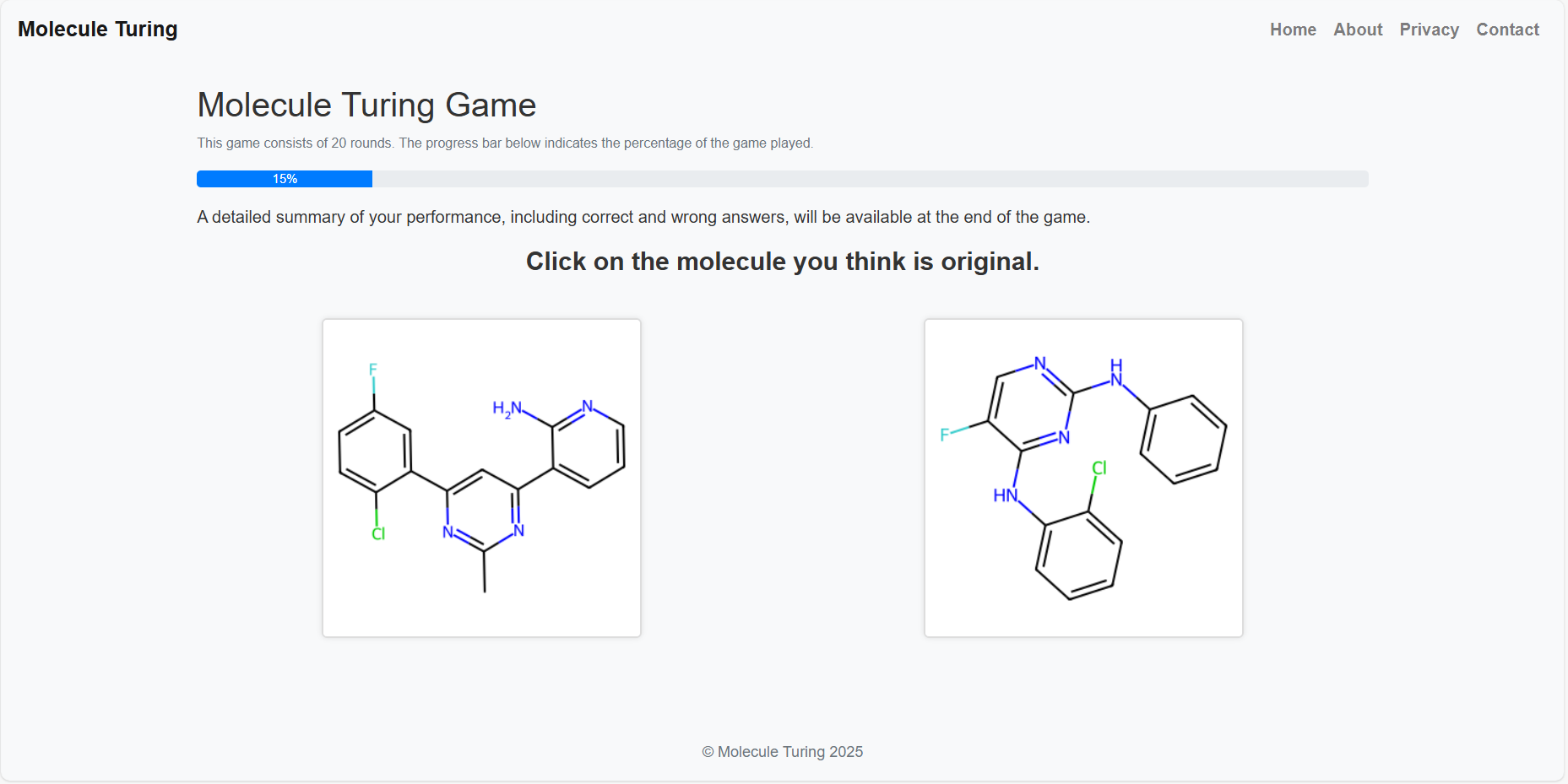} 
  \caption{{\bf Web user interface for the Turing-like test experiment.} At each round, 
  tThe web page presents two molecules with same molecular formula, one generated and one original, and the user has to guess which one is the original one. The user interface also shows a progress bar and additional information.}
  \label{fig:turing_test_web}
\end{figure}

\end{document}